\documentclass[10pt,twocolumn,letterpaper]{article}

\usepackage{cvpr}              

\usepackage{bm}
\usepackage{multirow}
\usepackage{graphicx}
\usepackage{siunitx}
\usepackage{makecell}
\usepackage{xcolor}
\usepackage[table]{xcolor}

\makeatletter
\@ifundefined{Cross}{}{}
\makeatother
\usepackage{marvosym}

\definecolor{cvprblue}{rgb}{0.21,0.49,0.74}
\definecolor{pinkcolor}{RGB}{230, 0, 115}
\definecolor{shadecolor}{gray}{0.9} 
\usepackage[pagebackref,breaklinks,colorlinks,allcolors=cvprblue]{hyperref}

\hypersetup{
    colorlinks=true,
    urlcolor=pinkcolor,
}

\def\paperID{} 
\def\confName{CVPR}
\def\confYear{2026}

\title{\raisebox{-0.3\height}{\includegraphics[height=1.8em]{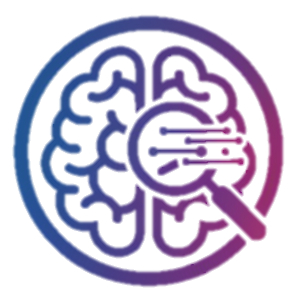}}\textbf{MemFlow: Flowing Adaptive Memory for Consistent and Efficient \\ Long Video Narratives}
\vspace{-0.9em}
}

\author{
Sihui Ji\textsuperscript{$1,\dagger$} \quad
Xi Chen\textsuperscript{1} \quad
Shuai Yang\textsuperscript{3} \quad
Xin Tao\textsuperscript{2} \\
Pengfei Wan\textsuperscript{2} \quad
Hengshuang Zhao\textsuperscript{1 \Letter}\\[0.3em]
{\textsuperscript{1}HKU \quad
\textsuperscript{2}Kling Team, Kuaishou Technology \quad
\textsuperscript{3}HKUST(GZ)
} \\[0.3em]
{\url{https://github.com/KlingTeam/MemFlow}}
}

\newcommand\nnfootnote[1]{%
  \begin{NoHyper}
  \renewcommand\thefootnote{}\footnote{#1}%
  \addtocounter{footnote}{-1}%
  \end{NoHyper}
}

\newcommand{\method}{\textsc{MemFlow}\xspace}
\newcommand{\bank}{Narrative Adaptive Memory\xspace}
\newcommand{\route}{Sparse Memory Activation\xspace}

\begin{document}
\twocolumn[{
\renewcommand\twocolumn[1][]{#1}
\maketitle
\vspace{-20pt}
\label{fig:teaser}
\begin{center}
    \vspace{-8pt}
    \includegraphics[width=0.93 \linewidth]{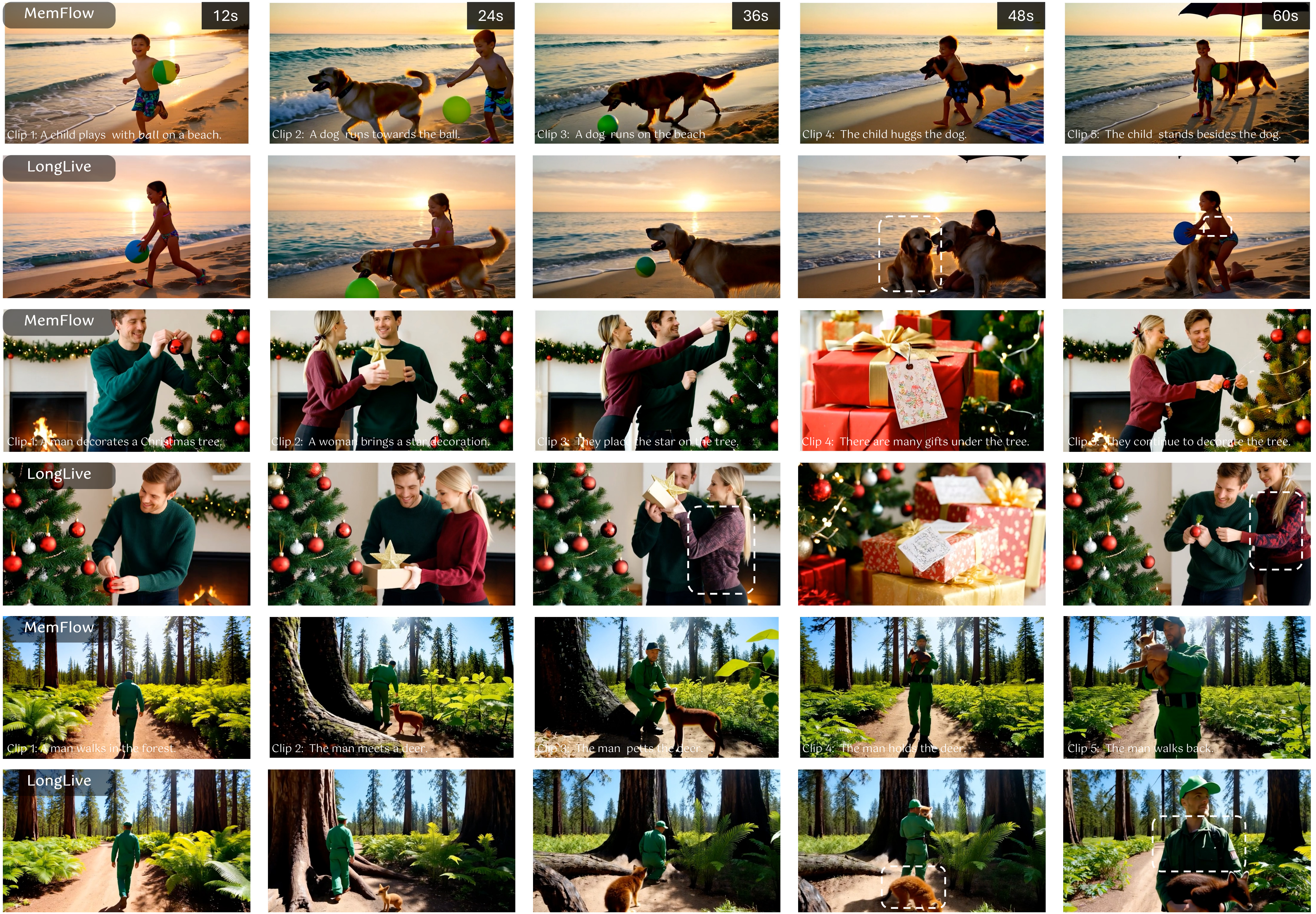}
    \vspace{-4pt}
    \captionsetup{type=figure}
    \caption{
    Existing streaming interactive text-to-video models such as LongLive~\cite{yang2025longlive} often fail to maintain consistency after prompt switching~(suffering from redundant subjects or inter-clip inconsistency).  \method addresses this by maintaining dynamic memory for long-term consistency, enabling narrative coherence even if new subjects appear or scenario switches.
    }
\end{center}
}]

\nnfootnote{
\hspace{-2em}$\dagger$ {This work was conducted during the author's internship at Kling Team, Kuaishou Technology.} \\
\hspace{2em}\Letter~{Corresponding Author.}
}

\vspace{-5pt}
\begin{abstract}

The core challenge for streaming video generation is maintaining content consistency over long context, which poses high requirement for the memory design. 
%
Most existing solutions maintain the memory by compressing historical frames with predefined strategies. 
However, different to-generate video chunks should refer to different historical cues, which is hard to satisfy with fixed strategies.
In this work, we propose \method to address this problem. 
Specifically, before generating the coming chunk, we dynamically update the memory bank by retrieving the most relevant historical frames with the text prompt of this chunk. This design not only accurately sources the context needed to maintain visual consistency, but also ensures semantic coherence even as new events unfold or scenes transition.
%
In addition, during generation, we only activate the most relevant tokens in the memory bank for each query in the attention layers, which effectively guarantees the generation efficiency. 
In this way, \method achieves outstanding long-context consistency with negligible computation burden~(7.9\% speed reduction compared with the memory-free baseline) and keeps the compatibility with any streaming video generation model with KV cache. 
\vspace{-3pt}
\end{abstract}

\vspace{-12pt}
\section{Introduction}
\label{sec:intro}
Video generation has attained remarkable quality~\cite{yang2024cogvideox, kling, kong2024hunyuanvideo, wan2025wan, sora}, making its extension to long durations critical for advancing creative and cinematic applications. While Diffusion Transformer~(DiT) models~\cite{dit} leverage bidirectional attention to capture complex spatiotemporal dependencies, their inherent computational costs and GPU memory limits constrain them to short video generation. Autoregressive~(AR) diffusion models~\cite{magi-1, causvid, self-forcing, skyreels-v2} offer a promising alternative by decomposing long videos into sequential clips, which alleviates the computation bottleneck through a reduced attention window.

Interactive video generation has emerged as a crucial application for enabling users to guide narratives with
streaming prompt inputs. Most existing works conduct chunk-level autoregressive generation, where new video segments are streamingly generated based on previously generated content and newly-provided text prompts. This interactive paradigm with dynamic prompt transitions allows the introduction of new elements and scene switches across extended temporal horizons.
However, it also poses difficulties in effectively preserving memory for long-range content consistency due to complex inter-clip dependencies.
First, since different to-generate video chunks should refer to different historical cues, the memory is required to adaptively provide relevant context according to streaming prompts; 
Second, the capacity of stored memory must be highly constrained, a necessity dictated by both the hardware limits of GPU memory and the demands of generation efficiency.

While the necessity of such adaptive and efficient memory module is evident, many existing approaches have been overly simplistic, failing to fully address the dual challenges outlined above.
They preserve memory in predefined paradigms, some only employ the first video chunk as memory sink~\cite{yang2025longlive}, some attempt to store more historical frames through fixed compression schemes~\cite{framepack, far,xiao2025captain}, some try to bake context implicitly with trainable memory modules~\cite{jiang2025lovic, test-time-training, hong2024slowfast}. 
However, those rigid strategies struggle to dynamically provide historical content corresponding to different prompt inputs, especially for new element emergence or scenario switches in prompt transitions.

Thus, we innovatively design \bank~(NAM), a memory mechanism that adaptively retrieves relevant historical content for interactive streaming video generation.
Specifically, we introduce a memory bank aggregating historical visual token~(KV cache) from streamingly generated chunks. During the sequential generation of each chunk, we first retrieve the context which aligns with the current prompt most, by calculating the attention score between textual token of the prompt and visual token from memory. The context frame with higher score is considered to be semantically relevant with current chunk generation, and will be integrated to update the memory along with a condensed representation of the immediately preceding chunk.
This design enables the current chunk to utilize historical cues, which have truly relevant content with the new prompt.
Our NAM is effective in preserving narrative coherence even if new event happens or scenario switches, which is hard to satisfy with fixed memory strategies.

However, the introduction of memory inevitably brings an extra computation burden, which hinders real-time generation. 
Thus, we propose Sparse Memory Activation~(SMA), which strategically activates only the most relevant tokens in attention layers according to the attention scores calculated from query~(current chunk) and key~(context in memory) by top-$k$ selection. Subsequent attention is then applied within these selected tokens, which effectively accelerates inference by reducing computation cost while preserving quality. 

In this way, our \method effectively maintains contextual consistency over long durations and adeptly balances the memory–efficiency trade-off. It achieves state-of-the-art quality for interactive video generation with only 7.9\% speed reduction compared with memory-free baseline. Our framework sustains 18.7 FPS on a single NVIDIA H100, 
demonstrating a clear advantage in producing narrative-coherent, long-term consistent videos with complex character and scene switching.

\section{Related Works}
\label{sec:related_work}

\begin{figure*}[t]
    \centering
    \includegraphics[width=1.0\linewidth]{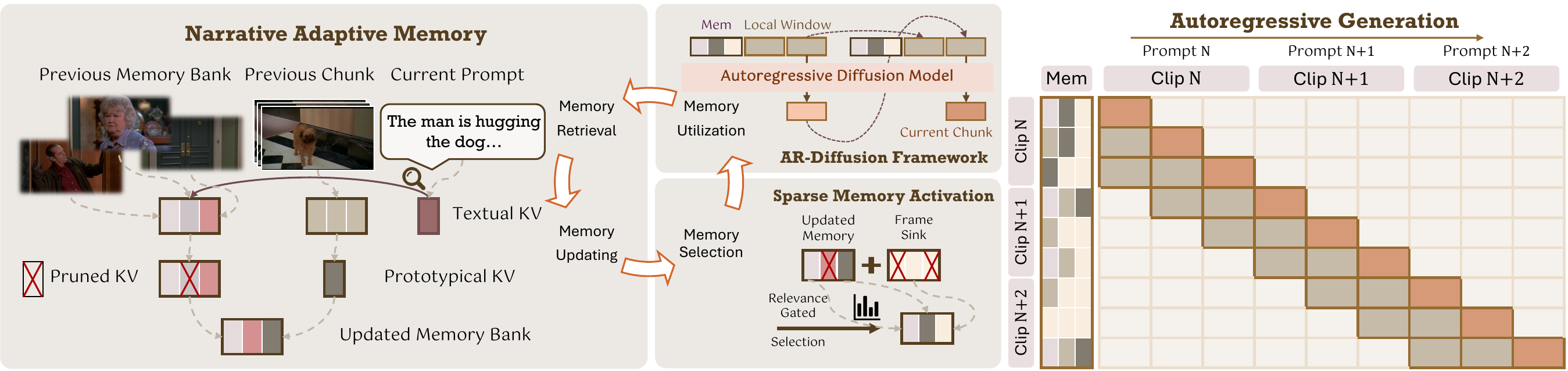}
    \vspace{-10pt}
    \caption{\textbf{Overall Framework of \method,} which is designed for interactive long video generation with both long-term consistency and efficiency. 
    In autoregressive generation, we conduct memory retrieval, updating, and selection sequentially before finally use it for synthesizing the incoming video chunk~(Sec~\ref{sec:framework}). Given the current prompt and KV cache memory bank, we first use textual token from the prompt to query the memory for retrieving semantically aligned KV cache. After adding the prototypical KV cache of previous chunk to inject the latest context,  the memory is updated for current chunk generation~(Sec~\ref{sec:nwm}). The updated memory is then combined with ''Frame Sink''-the KV cache from the first chunk-for the following Sparse Memory Activation. SMA filters the most relevant context for attention computation, improving the generation efficiency without sacrificing visual quality.~(Sec~\ref{sec:sma}).
    }
    \label{fig: pipeline}
    \vspace{-10pt}
\end{figure*}

\noindent \textbf{Long Video Generation.}
Prior efforts to extend video generation to longer durations can be broadly categorized into three approaches. {Autoregressive-diffusion hybrid approaches} generate long videos by iteratively predicting frames~\cite{causvid, self-forcing, skyreels-v2, magi-1, framepack, yang2025longlive}. 
Diffusion-Forcing~\cite{chen2025diffusionforcing} mitigates error propagation by adjusting denoising schedules. 
CausVid~\cite{causvid} distills bidirectional models into efficient few-step causal models, with Self Forcing~\cite{self-forcing} further addressing the train–test gap. MAGI-1~\cite{magi-1} and SkyReels-V2~\cite{skyreels-v2} successfully scale up this AR-diffusion paradigm.
{Multistage methods}~\cite{xiang2025macro, zhuang2024vlogger, huang2025filmaster, xiao2025captain} decompose a long video into multiple clips to be generated separately. They either first synthesize a sequence of coherent keyframes followed by video infilling for each clip~\cite{zhou2024storydiffusion, iclora, xiao2025captain}, or draft sequential prompts and use a T2V model to synthesize individual segments~\cite{zhao2025moviedreamer, long2024videostudio}. A fundamental limitation of these approaches is the isolated nature of clip generation, which often leads to a lack of temporal coherence over long horizons.
The third category applies {efficient architectures} to manage computational costs. TTTVideo~\cite{test-time-training} and LaCT~\cite{zhang2025lact} learn context using neural networks with linear attention. TokensGen~\cite{ouyang2025tokensgen} represents video clips with condensed tokens.
Mixture of Contexts~\cite{cai2025mixture} dynamically selects relevant context for attention computation. These methods often sacrifice visual fidelity for efficiency.

\noindent \textbf{Memory Mechanisms in Video Generation.}
Effective memory mechanisms are crucial for maintaining consistency in long video generation.
{Action-guided video generation} often relies on geometric and spatial dependencies~\cite{zhai2025stargen, chen2025learning, xiao2025worldmem}. Worldmem~\cite{xiao2025worldmem} and Context as memory~\cite{yu2025context} conduct memory retrieval based on Field of View~(FOV) overlap between conditioned camera poses. VMem~\cite{li2025vmem} introduces Surfel-Indexed View Memory for efficient retrieval by indexing past views with 3D surface elements. These methods, however, are highly dependent on spatial priors, thus lack generalizability. {General video generation} primarily maintains memory through context compression~\cite{framepack, far, hong2024slowfast}. FramePack~\cite{framepack} compresses input frames into a fixed-size context to manage memory and efficiency. FAR~\cite{far} and StreamingT2V~\cite{henschel2025streamingt2v} combine short- and long-term memory via multiscale compression and learnable modules, respectively. These methods often maintain memory without adaptive retrieval, making it challenging to build dynamic connections between relevant context and the currently generated clip.
\section{Method}
\label{sec:method}
~\method enhances long-video narrative consistency by incorporating a novel dynamic memory bank into a streaming video generation framework~(Sec \ref{sec:framework}). 
To dynamically recall relevant historical context according to current prompt, we first adopt~(1) \bank~(NAM) mechanism for memory retrieval and updating~(Sec \ref{sec:nwm}); then~(2) \route~(SMA) is employed for memory selection to address memory-efficiency trade-off~(Sec \ref{sec:sma}). The refreshed memory is subsequently utilized by AR-diffusion model to synthesize the current video chunk, after which this process continues to roll out over extended temporal horizon.
The entire framework is trained end-to-end using a streaming long-tuning strategy, enabling the model to learn how to effectively manage its memory during long-duration rollouts. 
Figure~\ref{fig: pipeline} provides a high-level illustration of our framework.

\subsection{Overall Framework}
\label{sec:framework}
\textbf{Baseline.} Our work builds upon a hybrid autoregressive-diffusion framework that integrates autoregressive chunk-wise video generation with denoising diffusion~\citep{causvid, magi-1, self-forcing}.
At each generation iteration, the model produces a chunk of $T$ frames, conditioned on the $n$ immediately preceding frames. 
This autoregressive process naturally produces Key-Value (KV) cache from previous iterations, which is leveraged as the foundational structure for our memory bank. 
This design allows us to store historical context efficiently without incurring extra computational overhead. 
In a standard setup, the autoregressive attention mechanism operates over ($n+T$) local frames~(the last $n$ preceding frames and current $T$ generated frames). 
By integrating our memory bank containing $B$ frames, the attention operation is extended to cover ($n+B+T$) frames, seamlessly blending short-term dependencies and long-term memory.

\noindent \textbf{Training Mechanism.} We train our memory-augmented AR-diffusion model using a distillation-based approach, specifically adopting the Self-Forcing~\cite{self-forcing} paradigm.
Specifically, we adopt Distribution Matching Distillation~(DMD) loss~\citep{dmd} that minimizes the gap between the student and teacher generator's output distribution, to
distill a pretrained bidirectional model into a few-step causal model.
To equip the model with long-context capabilities, we employ a streaming long-tuning strategy~\cite{yang2025longlive}.
During this phase, the generator samples a short video clip~(e.g., 5s) in each round conditioned on previous clips, and the teacher provides reliable supervision on this newly generated short clip via DMD. 
We can repeat this rolling extension for generating long sequences until the video reaches a preset maximum length, with supervision applied throughout the entire rollout.
Crucially, we integrate our memory mechanism (NAM and SMA) into this tuning process: employing NAM in the
streaming tuning allows the model to learn \textit{how} to retrieve relevant history from self-generated frames during training, aligning training with inference and improving long-range consistency; while SMA mitigates the computational overhead introduced by memory, incurring only 7.9\% efficiency loss compared to the memory-free baseline.

\vspace{-2pt}
\subsection{\bank~(NAM)}
\label{sec:nwm}
We first formulate the components of our memory bank in NAM. At each iteration, a new chunk is generated autoregressively, during which it is processed by the DiT to produce key-value~(KV) representations ${\{\bm{K}_m^l, \bm{V}_m^l\}_{l=1}^{L}}$ at each transformer layer $l$, where $m$ denotes the index of chunk generation iteration.
At the beginning of next iteration, the memory is updated as ${\{\bm{K}_m^{l'}, \bm{V}_m^{l'}\}_{l=1}^{L}}$ for subsequent computation. 
Our memory mechanism aims to provide content-aligned context for incoming generation, which necessitates its ability to retrieve history relevant to incoming prompt and incorporate the most recently generated content for updating. To avoid excessive expansion of the memory bank as generation proceeds, we introduce two synergistic techniques:~(i) Semantic Retrieval, which retrieve most informative context based on the cross-attention relevance between textual queries and visual keys, and~(ii) Redundant Removal, which leverages temporal redundancy to select the KV feature of first latent frame as prototype for the entire local chunk.



\noindent\textbf{Semantic Retrieval.}
During generation, each transformer layer $l$ produces key–value representations for the current chunk while attending across the present sequence, the KV cache in local window, and the global memory bank.
The retrieval criterion is derived from cross-attention scores between the textual tokens as query and visual tokens from KV cache as key, 
which has proven effective in prior works~\cite{chen2024image, wang2025adaretake, yang2025streammem} of large vision-language models. In our design, the textual tokens are computed from the prompt of chunk to be generated, thus the visual tokens with higher scores are semantically-aligned with this chunk. By retrieving those KV cache in the memory bank, we expect the model to attend to content-relevant visual features.

Let $\bm{Q}^l_{\text{text}} \in \mathbb{R}^{d}$ be the textual query of the current text prompt at layer $l$ . For each of the $b$ frames stored in the memory bank, represented by its key of KV cache $\bm{K}_{m,i}^l \in \mathbb{R}^{n \times d}$ where $i=1, \dots, b$, we compute a semantic relevance score, $\mathcal{S}_{m,i}^l$:
\begin{equation}
\label{eq:relevance_score_new}
\mathcal{S}_{m,i}^l = \operatorname{Aggregate}\left( \operatorname{Softmax}\!\left(
        \frac{\bm{Q}^l_{\text{text}} (\bm{K}_{m,i}^l)^\top}{\sqrt{d}}
    \right)\right),
\end{equation}
where the $\operatorname{Softmax}(\cdot)$ computes attention weights, and $\operatorname{Aggregate}(\cdot)$ is mean pooling here to produce a scalar score $\mathcal{S}_{m,i}^l \in \mathbb{R}$. Then we can identify the top-$k$ most semantically aligned frames to be retained.

\noindent\textbf{Redundant Removal.}
Following Semantic Retrieval, the immediately preceding chunk is consolidated into a representative prototype before being integrated into the memory. Instead of adopting computationally intensive context-merging techniques~\cite{he2024ma, yang2025streammem, zhang2025beyond} that rely on importance weighting, we propose a highly efficient heuristic. We leverage the high temporal redundancy inherent in short video chunks, where visual information exhibits strong similarity across consecutive frames. We posit that a single frame is sufficient to encapsulate the core visual content of the entire chunk. Therefore, we simply select the KV pair of the first frame from the preceding chunk to serve as its compact prototype.
The updated memory bank $\{\bm{K}_m^{l\prime}, \bm{V}_m^{l\prime}\}$ is then formed by concatenating the selected historical frames with the newly consolidated local prototype. The two strategies ensure that the memory is semantically relevant and real-time updated, enabling the model to build long-term and short-term dependencies crucial for narrative coherence.




\subsection{Sparse Memory Activation~(SMA)}
\label{sec:sma}

\begin{table*}[t]
\centering
\scriptsize
\caption{\textbf{Quantitative comparison under multi-prompt 60-second setting} with representitive long video generation models, where SkyReels-V2~\cite{skyreels-v2}, Self Forcing~\cite{self-forcing} and FramePack~\cite{framepack} are adapted for the task by directly switching prompts. All scores are measured over the whole sequence, except for the CLIP score, which is computed at intervals aligned with the prompt switching.}
\resizebox{\linewidth}{!}{
\begin{tabular}{l c c c *{6}{c}}
\toprule
\multirow{2}{*}{\textbf{Method}} &
\multicolumn{1}{c}{\multirow{2}{*}{\makecell{Quality\\Score $\uparrow$}}} &
\multicolumn{1}{c}{\multirow{2}{*}{\makecell{Consistency\\Score $\uparrow$}}} &
\multicolumn{1}{c}{\multirow{2}{*}{\makecell{Aesthetic\\Score $\uparrow$}}} &
\multicolumn{6}{c}{CLIP Score $\uparrow$} \\
\cmidrule(lr){5-10} 
& & & & \multicolumn{1}{c}{0--10\,s} & \multicolumn{1}{c}{10--20\,s} &
\multicolumn{1}{c}{20--30\,s} & \multicolumn{1}{c}{30--40\,s} &
\multicolumn{1}{c}{40--50\,s} & \multicolumn{1}{c}{50--60\,s} \\
\midrule
SkyReels-V2~\citep{skyreels-v2}   &  81.55 & {94.72} & {56.83} & 25.31 & 23.40 & 22.50 & 21.62 & 21.67 & 20.91\\ 
Self Forcing~\citep{self-forcing}  & 83.94 & {95.74} & {58.45}& 26.24 & \underline{24.87} & 23.46 & 21.92 & 22.05 & 21.07 \\ 
LongLive~\citep{yang2025longlive}      & 84.28 & {96.05} & \underline{59.89} & \textbf{26.63} & \textbf{25.77} & \textbf{24.65} & \underline{23.99} & \underline{24.52} & \underline{24.11}\\ %
FramePack~\citep{self-forcing}  & \underline{84.40} & \textbf{96.77} & 59.44 & \underline{26.51} & 22.60 & 22.18 & 21.53 & 21.98 & 21.62 \\ 
\textbf{\method}      & \textbf{85.02} & \underline{96.60} & \textbf{61.07}& {26.31} & {24.70} & \underline{23.94} & \textbf{24.13} & \textbf{24.90} & \textbf{24.22} \\ 
\bottomrule
\end{tabular}}
\label{tab:comp_interactive}
\vspace{-10pt}
\end{table*}

Directly extending local context window to incorporate a memory bank introduces computational burden, as attention complexity scales with context size. While rigidly compressing the context can improve efficiency, it often compromises memory quality, as critical historical cues may be discarded indiscriminately. To address this memory–efficiency trade-off, we introduce \route, a relevance-gated memory selection technique for dynamic memory pruning before attention computation.

Our approach operates on the principle of selective attention, where query token from the current video chunk attends only to a subset of the most relevant historical frames in memory. Formally, we first partition key~($\bm{K}_m^l$) and value~($\bm{V}_m^l$) of the memory bank into $b$ frames. We then compute a compact descriptor for both the query~($\bm{Q_{\text{vis}}}^l$) from current chunk and key of each frame using mean pooling over the token dimension, which is highly sufficient and expressive for generation tasks and has demonstrated by prior works~\cite{cai2025mixture}.
This yields a single query descriptor, $\bar{\bm{q}}_{\text{vis}}\in \mathbb{R}^{1 \times d}$, and a set of frame-wise key descriptors, $\{\bar{\bm{k}}_{j}\}_{j=1}^{b}\in \mathbb{R}^{1 \times d}$, the chunk index $m$ is omitted here for simplicity.
The relevance between the current query and frame-wise key in memory is then determined by the inner product of their respective descriptors:
\begin{equation}
\label{eq:relevance_score}
s_j = \bar{\bm{q}}^\top_{\text{vis}} \bar{\bm{k}}_j, \quad \text{for} \quad j = 1, \dots, b
\end{equation}
Based on these relevance scores, we identify the set of indices $\mathcal{I}_k$ corresponding to the top-$k$ most relevant frames:
\begin{equation}
\label{eq:topk_indices_argmax}
\mathcal{I}_k = \underset{I \subseteq \{1, \dots, b\}, |I|=k}{\operatorname{arg\,max}} \sum_{j \in I} s_j
\end{equation}
This formulation selects the subset of indices $I$ of size $k$ that maximizes the sum of relevance scores.
Finally, the attention computation for query $\bm{Q_{\text{vis}}}^l$ is restricted to the key-value pairs belonging to the selected top-$k$ chunks:
\begin{equation}
\label{eq:sparse_attention}
\operatorname{Attn}(\bm{Q_{\text{vis}}}^l, \bm{K}_m^l, \bm{V}_m^l) \approx \operatorname{Attn}(\bm{Q_{\text{vis}}}^l, \bm{K}_{m, \mathcal{I}_k}^l, \bm{V}_{m, \mathcal{I}_k}^l)
\end{equation}
where $\bm{K}_{m, \mathcal{I}_k}^l$ and $\bm{V}_{m, \mathcal{I}_k}^l$ are the concatenated key and value tensors from the chunks indexed by the set $\mathcal{I}_k$.

By activating part of the memory bank, SMA reduces computational latency while retaining the most pertinent historical information. This strategy enables the model to selectively recall the right context at the right time, thereby preserving long-range dependencies and narrative coherence. Moreover, by implicitly filtering out less relevant or potentially erroneous information from the history, our approach mitigates error accumulation. This allows \method to achieve both robust memorization and computational efficiency, ensuring the generation of coherent long videos without the degradation of visual quality over time.

\section{Experiment}
\label{sec:experiment}
\begin{table*}[t]
  \caption{
    \textbf{Quantitative comparisons under single-prompt 5-second setting} with representative open-source video generation models of similar parameter sizes and resolutions. Our \method performs on par with other models on overall quality and has a clear advantage in semantic score attributed to textual retrieval-based memory. {\raggedright$^{\dagger}$ denotes the scores reproduced by us.}
  }
  \label{tab:comp_short}
  \vspace{-5pt}
  \centering
\begin{tabular}{lccccccc}
  \toprule
  \multirow{2}{*}{Model} & \multirow{2}{*}{\#Params} & \multirow{2}{*}{Resolution} & \multirow{2}{*}{\makecell{Throughput\\(FPS) $\uparrow$}} & \multicolumn{3}{c}{Evaluation scores $\uparrow$}\\
  \cmidrule(lr){5-7}
   &  &  &  & Total & Quality & Semantic \\
  \midrule
  \rowcolor{shadecolor}
  \multicolumn{7}{l}{\textit{Diffusion models}}\\
  LTX-Video~\citep{HaCohen2024LTXVideo}      & 1.9B & $768{\times}512$ & 8.98          & 80.00 & 82.30 & 70.79 \\
  Wan2.1~\citep{wan2025wan}                         & 1.3B & $832{\times}480$ & 0.78          & 84.26 & 85.30 & 80.09 \\
  \midrule
  \rowcolor{shadecolor}
  \multicolumn{7}{l}{\textit{Autoregressive models}}\\
  SkyReels-V2~\citep{skyreels-v2}            & 1.3B & $960{\times}540$ & 0.49          & 82.67 & 84.70 & 74.53 \\
  MAGI-1~\citep{magi-1}                      & 4.5B & $832{\times}480$ & 0.19          & 79.18 & 82.04 & 67.74 \\
  CausVid~\citep{causvid}                    & 1.3B & $832{\times}480$ & 17.0 & 81.20 & 84.05 & 69.80 \\
  NOVA~\citep{deng2024nova}                  & 0.6B & $768{\times}480$ & 0.88          & 80.12 & 80.39 & 79.05 \\
  Pyramid Flow~\citep{pyramid-flow}          & 2B   & $640{\times}384$ & 6.7           & 81.72 & 84.74 & 69.62 \\
  Self Forcing, chunk-wise~\citep{self-forcing} & 1.3B & $832{\times}480$ & 17.0    & 84.31 & 85.07 & \underline{81.28} \\
  Self Forcing, frame-wise~\citep{self-forcing} & 1.3B & $832{\times}480$ & 8.9           & 84.26 & 85.25 & 80.30 \\
  LongLive~\cite{yang2025longlive}                                     & 1.3B & $832{\times}480$ & \textbf{20.3}$^{\dagger}$          & \underline{84.87} & \textbf{86.97} & 76.47 \\
  \textbf{\method}                                     & 1.3B & $832{\times}480$ & \underline{18.7}         & \textbf{85.14} & \underline{85.95} & \textbf{81.90} \\
  \bottomrule
\end{tabular}
\vspace{-5pt}
\end{table*}

\begin{figure*}[t]
    \centering
    \includegraphics[width=1.0\linewidth]{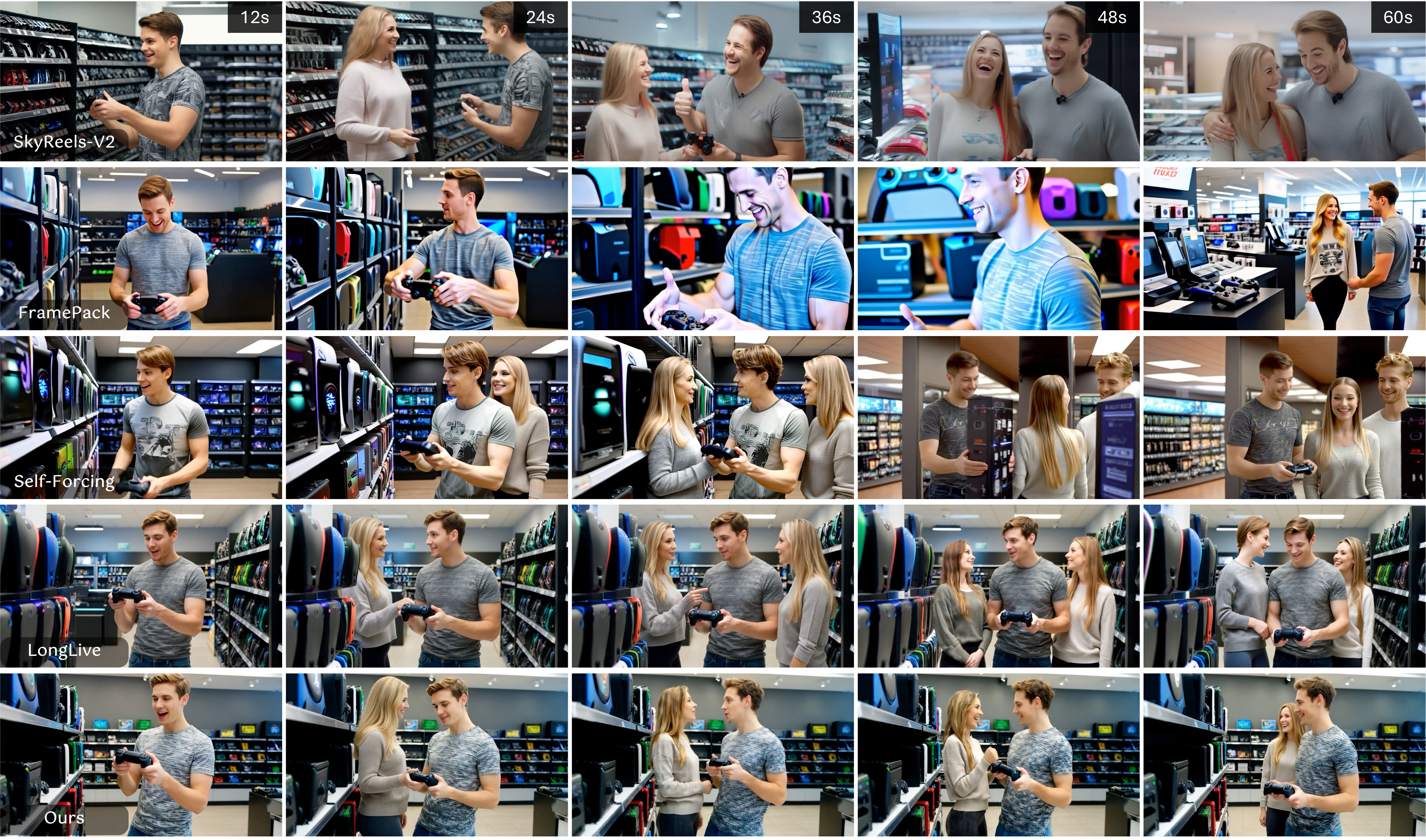}
    \caption{\textbf{Qualitative comparisons under multi-prompt 60-second setting} with representative long video generation models, where \method outperforms other alternatives in narrative coherence and subject consistency, without drifting or duplicated characters.}
    \label{fig:comp_interative}
    \vspace{-10pt}
\end{figure*}

\begin{figure*}[t]
    \centering
    \includegraphics[width=1.0\linewidth]{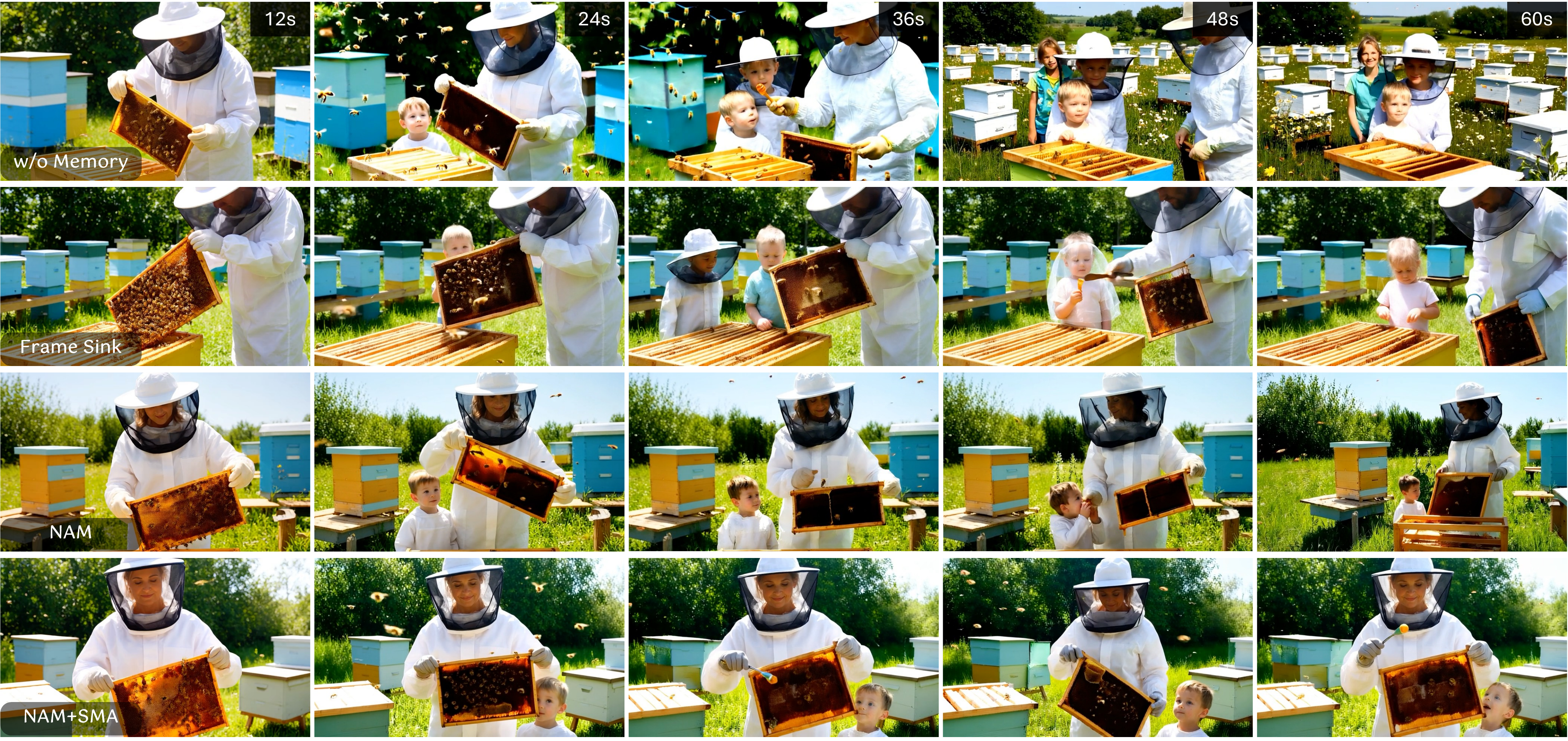}
    \caption{
    \textbf{Qualitative analysis of different memory mechanisms} under multi-prompt 60-second setting. ``w/o Memory'' means only attending to the local attention window, ``Frame Sink'' refers to keeping KV cache from the first chunk as memory, ``NAM'' adopts the whole memory bank without filtering, and ``NAM+SMA'' is our full model which compresses memory by relevance-gated selection.}
    \label{fig:abla_interative}
    \vspace{-12pt}
\end{figure*}

\subsection{Implementation Details}
\label{implementation}
We build \method on Wan2.1-T2V-1.3B~\citep{wan2025wan}, following the training and inference pipeline from LongLive~\cite{yang2025longlive}, while enabling our memory bank and sparse activation.
We perform Self Forcing~\cite{self-forcing} DMD pipeline with streaming long tuning on a 60s sequence by using the switch-prompt dataset from LongLive constructed by Qwen2-72B-Instruct~\citep{qwen22024}. We conduct streaming long tuning equipped with the memory bank for 3000 steps.
During training, each iteration continues the model’s own rollout by generating the next 5s video clip until a maximum length of 60s is reached.




\subsection{Comparisons for Multi-prompt Generation}
Since our memory mechanism is designed for interactive long-form videos with multiple prompts, we first compare \method's abilities with representative long video generation models. 
For fair comparison, we adapt existing methods - including SkyReels-V2~\cite{skyreels-v2}, FramePack~\cite{framepack}, and Self Forcing~\cite{self-forcing} - for multi-prompt video generation by switching prompts during the autoregressive synthesis. Note that LongLive~\cite{yang2025longlive} inherently supports generating videos with interactive instructions. Following LongLive~\cite{yang2025longlive}, we customize 100 groups of narrative scripts, with each consisting of 6 successive 10-second prompts for a total of 100 videos lasting for 60 seconds. 

We use metrics from VBench-Long~\cite{huang2024vbench++} for assessing visual quality of all generated videos, among which the two dimensions of consistency and aesthetic are highlighted for comparison of long-range consistency in subjects, background and visual aesthetics.
Table~\ref{tab:comp_interactive} shows that \method achieves the best quality score among all methods, verifying its comprehensive competitiveness in perceptual quality. In terms of consistency score, our method outperforms all other models except Framepack~\cite{framepack}, with Framepack tending to synthesize videos with reduced inter-frame dynamics, which definitely shows superiority in ``consistency''. Thus the result can still demonstrate our superior performance in global consistency with the specially designed memory. Our advantage in aesthetic score also highlights the effectiveness of our method in mitigating error accumulation during long rollouts.



For text alignment with interactive prompts, videos are segmented according to prompt boundaries for evaluating clip-wise semantic adherence. The CLIP score~\citep{radford2021learningtransferablevisualmodels} between each video clip and its corresponding text is calculated at 10-second intervals. The results demonstrate outstanding prompt adherence and narrative coherence, particularly when the video is extrapolated to longer durations, owing to the model’s ability to establish long-term contextual associations. Such superiority is further corroborated by the qualitative results in Figure~\ref{fig:comp_interative}. Our \bank successfully links description in the prompt, ``a woman in a casual sweater'', with the exact person in previous frames, thus maintaining the subject consistency.
While our baseline, LongLive~\cite{yang2025longlive}, fails to make connections between visual cues and semantic instructions, therefore continuously introduces new characters after prompt switching, exhibiting less temporal coherence and prompt compliance. Other approaches exhibit more severe error accumulation, with subject inconsistency in SkyReels-V2~\cite{skyreels-v2} and color drifting in FramePack. 
Self Forcing~\cite{self-forcing} also suffers from similar problems with LongLive, showing misalignment between prompt scripts and narrative progression across clips as characters are repeatedly introduced into the ongoing scene. Additionally, We conducted a user study with 20 participants to compare our method against the aforementioned models, and report the result in supplementary material. It includes human preference rates in visual quality, instruction following, and global consistency, further supporting the effectiveness of our approach.
In terms of speed, \method is more than 38$\times$ faster than SkyReels-v2~\cite{skyreels-v2}, slightly faster than Self Forcing~\cite{self-forcing}, while slightly slower than LongLive~\cite{yang2025longlive} due to memory updating and activation.


\subsection{Comparisons for Single-prompt Generation}
Although not specifically trained for single-prompt generation, our model shows superior performance compared to state-of-the-art models for durations of 5s and 30s on VBench~\cite{huang2023vbench} official prompt set.

\noindent \textbf{Short Video Generation.}
We evaluate \method’s short-video generation and compare it with relevant open-source models of similar scale, including LTX-Video~\citep{HaCohen2024LTXVideo}, Wan2.1~\citep{wan2025wan}, SkyReels-V2~\citep{skyreels-v2}, MAGI-1~\citep{magi-1}, CausVid~\citep{causvid}, NOVA~\citep{deng2024nova}, Pyramid Flow~\citep{pyramid-flow}, Self Forcing~\citep{self-forcing}, and LongLive~\citep{self-forcing}. 
For 5-second videos, \method achieves strong overall quality with the highest total score, compared with state-of-the-art models as in Table~\ref{tab:comp_short}.
By retrieving relevant context from \bank via prompt query, our model surpasses all other models in semantic score.
Due to computation cost in memory updating and activation, our \method sacrifices inference speed by 8.6\% compared with LongLive, but still outperforms other methods with 18.7 FPS for real-time inference. The results also show that our framework does not degrade short-clip generation capability.

\begin{table*}[t]
\centering
\scriptsize
\caption{\textbf{Quantitative analysis of different memory mechanisms} under multi-prompt 60-second setting. ``w/o Memory'' means only attending to the local attention window, ``Frame Sink'' refers to keeping KV cache from the first chunk as memory~\cite{yang2025longlive}, ``NAM'' adopts the whole memory bank without filtering, and ``NAM+SMA'' is our full model which compresses memory by relevance-gated selection.}
\resizebox{\linewidth}{!}{
\begin{tabular}{l c c c *{6}{c}}
\toprule
\label{tab:abla_interative}
\multirow{2}{*}{\makecell[l]{\textbf{Memory}\\\textbf{Mechanism}}} &
\multicolumn{1}{c}{\multirow{2}{*}{\makecell{Subject\\Consistency $\uparrow$}}} &
\multicolumn{1}{c}{\multirow{2}{*}{\makecell{Background\\Consistency $\uparrow$}}} &
\multicolumn{1}{c}{\multirow{2}{*}{\makecell{Throughput\\(FPS) $\uparrow$}}} &
\multicolumn{6}{c}{CLIP Score $\uparrow$} \\
\cmidrule(lr){5-10} 
& & & & \multicolumn{1}{c}{0--10\,s} & \multicolumn{1}{c}{10--20\,s} &
\multicolumn{1}{c}{20--30\,s} & \multicolumn{1}{c}{30--40\,s} &
\multicolumn{1}{c}{40--50\,s} & \multicolumn{1}{c}{50--60\,s} \\
\midrule
w/o Memory   &  94.41 & {95.15} & \textbf{23.5} & 26.74 & 25.10 & 24.60 & \textbf{23.61} & 24.23 & 24.14 \\ 
Frame Sink~\citep{yang2025longlive}  & {97.66} & {96.20} & \underline{20.3}& \textbf{26.63} & \textbf{25.77} & \textbf{24.65} & {23.99} & {24.52} & {24.11}\\
NAM+SMA      & \underline{98.01} & \textbf{96.70} & {18.7} & {26.31} & {24.70} & {23.94} & \underline{24.13} & \underline{24.90} & \underline{24.22} \\ 
NAM      & \textbf{98.05} & \underline{96.57} & {17.6} & \underline{26.50} & \underline{25.30} & \underline{24.42} & \textbf{24.23} & \textbf{24.96} & \textbf{24.28} \\ 
\bottomrule
\end{tabular}}
\vspace{-5pt}
\end{table*}

\noindent\textbf{Long video generation.}
The superiority of our method becomes more pronounced in long-horizon single prompt generation for 30-second videos. We observe consistent improvements across quality and semantic metrics in Table~\ref{tab:comp_long}, leading to an outstanding overall performance than SkyReels-V2~\citep{skyreels-v2}, FramePack~\citep{framepack}, Self Forcing~\citep{self-forcing}, and LongLive~\citep{yang2025longlive}. 
It verifies that our \bank provides more semantic-consistent context for video generation over long duration compared with using local context window only~(SkyReels-V2, Self Forcing), context compression~(FramePack) or retaining the first chunk as memory~(LongLive). Moreover, the retrieval-based memory updating strategy interrupts the error propagation in memorization implicitly-only context with the highest level of semantic adherence is included for attention computation-thus alleviating the degradation of visual quality due to error accumulation over time. \method maintains an advantage in long video generation quality with comparable performance on efficiency.



\begin{table}[t]
\centering
\vspace{-5pt}
\caption{\textbf{Quantitative comparisons for single-prompt 30-second setting} with representative long video generation models, showing more pronounced superiority than 5-second setting on all metrics with efficiency comparable to state-of-the-art models.
}
\label{tab:comp_long}
\small
\setlength{\tabcolsep}{4pt}
\renewcommand{\arraystretch}{1}
\begin{tabular}{lcccc}
\toprule
\textbf{Model} & \makecell{Total\\Score $\uparrow$} & \makecell{Quality\\Score $\uparrow$} & \makecell{Semantic\\Score $\uparrow$} & \makecell{Throughput\\(FPS) $\uparrow$} \\
\midrule
SkyReels-V2~\cite{skyreels-v2}  & 75.29 & 80.77 & 53.37  & 0.49 \\
FramePack~\cite{framepack}    & 81.95 & 83.61 & 75.32  & 0.92 \\
Self Forcing~\cite{self-forcing} & 81.59 & 83.82 & 72.70  & 17.0 \\
LongLive~\cite{yang2025longlive}      & 83.52 & 85.44 & 75.82  & \textbf{20.3} \\
\textbf{\method}      & \textbf{84.51} & \textbf{85.92} & \textbf{78.87}  & \underline{18.7} \\
\bottomrule
\end{tabular}
\vspace{-10pt}
\end{table}

\begin{figure}[t]
    \centering
    \includegraphics[width=1.0\linewidth]{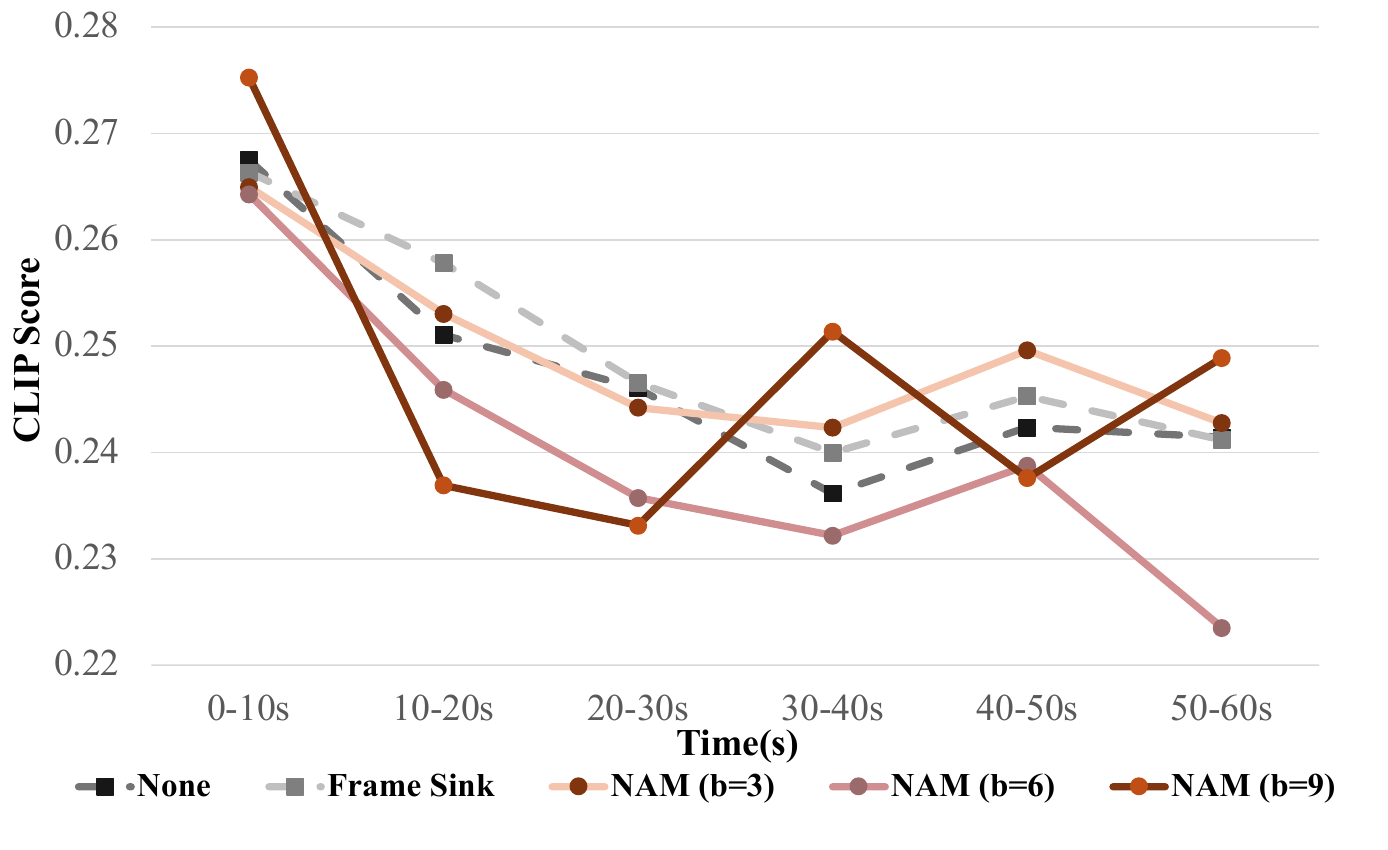}
    \vspace{-15pt}
    \caption{\textbf{Quantitative analysis of different memory capacity} under multi-prompt 60-second setting. ``w/o Memory'' means only attending to the local attention window, ``Frame Sink'' refers to keeping KV cache from the first chunk as memory~\cite{yang2025longlive}, ``NAM'' adopts the whole memory bank including $b$ latent frames.}
    \label{fig:abla_cap}
    \vspace{-20pt}
\end{figure}

\subsection{Ablation Studies}
We perform an ablation study on the core designs of our framework, \bank and \route, in a 60-second interactive multi-prompt video generation setting with five prompt switches.

\noindent\textbf{Memory Mechanism.}
In Table~\ref{tab:abla_interative}, we ablate different memory mechanism by comparing~(i) w/o memory: conditioned only on local context window;~(ii) Frame Sink: addtionally retain the KV cache from the first chunk;~(iii) \bank~(NAM): maintain our dynamic memory bank; and~(iiii) \bank and \route~(NAM+SMA): our full model.
Frame Sink is utilized by LongLive~\cite{yang2025longlive}, allowing direct comparison with its model. The configuration of w/o memory is also implemented on the baseline of LongLive by removing the sink latent frames. Table~\ref{tab:abla_interative} highlights the effectiveness of our NAM, which consistently outperforms others in maintaining temporal consistency and semantic coherence. The retrieval-based memory establishes intrinsic dependencies across contexts, enabling stable narrative transitions even under subject insertion or switching. With SMA, inference efficiency improves from 17.6 to 18.7 frames per second with minimal quality degradation. As shown in Figure~\ref{fig:abla_interative}, removing memory causes abrupt scene transitions, while Frame Sink preserves continuity only for initial subjects but collapses on later ones. In contrast, our model captures the relations between existing and emerging subjects, achieving superior semantic alignment on switched prompts, especially as the video extends beyond 30 seconds.


\noindent\textbf{Memory Capacity.}
In Figure~\ref{fig:abla_cap}, we ablate the impact of our two key components under the 60-second setting: the capacity of origin memory in NAM and activated memory after SMA. 
The \textit{left} panel analyzes NAM with varying capacities $b=\{3, 6, 9\}$ against two baselines ``w/o Memory'' and ``Frame Sink''. 
The results reveal that a larger memory capacity does not guarantee better performance. Notably, \texttt{NAM~(b=6)}  consistently underperforms the baseline, while \texttt{NAM~(b=9)} exhibits significant performance instability. We attribute this to an imbalance within the attention's receptive field: as memory capacity $b$ increases, the proportion of global context from memory significantly outweighs that of the local window. This over-reliance on global context can disrupt short-term narrative flow, leading to the observed fluctuations in CLIP score. Therefore, we select \texttt{NAM~(b=3)}, a capacity equivalent to half the size of our local context window, as it provides the most stable  balance between local and global context and effective enhancement in semantic coherence over baseline.


\section{Conclusion}
\label{sec:conclusion}
In this work, we introduce \method, a memory mechanism for equipping interactive long video generation with long-range consistency without severe efficiency degradation. For maintain long-term memory for narrative 
coherence, we design \bank for dynamic retrieval of semantic-aligned context via textual query.  We also introduce \route for balancing the memory-efficiency trade-off by relevance-gated memory filtering.  Our model achieves 18.7 FPS inference on a single NVIDIA H100 GPU, and supports interactive video generation while maintaining consistency, visual quality and narrative coherence even under complex narrative transitions and character switching.

{
    \small
    \bibliographystyle{ieeenat_fullname}
    \bibliography{main}
}


\end{document}